# Comparison Analysis of Traditional Machine Learning and Deep Learning Techniques for Data and Image Classification

EFSTATHIOS KARYPIDIS[1], STYLIANOS G. MOUSLECH[1], KASSIANI SKOULARIKI[2], ALEXANDROS GAZIS[1,*]

[1]Democritus University of Thrace, Department of Electrical and Computer Engineering
Xanthi, 67100
GREECE

[2]Democritus University of Thrace, Department of Production Engineering and Management
Xanthi, 67100
GREECE

Abstract — The purpose of the study is to analyse and compare the most common machine learning and deep learning techniques used for computer vision 2D object classification tasks. Firstly, we will present the theoretical background of the Bag of Visual words model and Deep Convolutional Neural Networks (DCNN). Secondly, we will implement a Bag of Visual Words model, the VGG16 CNN Architecture. Thirdly, we will present our custom and novice DCNN in which we test the aforementioned implementations on a modified version of the Belgium Traffic Sign dataset. Our results showcase the effects of hyperparameters on traditional machine learning and the advantage in terms of accuracy of DCNNs compared to classical machine learning methods. As our tests indicate, our proposed solution can achieve similar - and in some cases better - results than existing DCNNs architectures. Finally, the technical merit of this article lies in the presented computationally simpler DCNN architecture, which we believe can pave the way towards using more efficient architectures for basic tasks.

Keywords - Computer Vision, Multi-Class Classification, K-Means Clustering, Feature Vectors, Linear/Non-Linear Classifiers, Deep Learning, Convolutional Neural Networks, Data Augmentation, Feature Maps (Feature Learning), Transfer Learning

## 1. Introduction

IN our modern day and age, the amount of data provided daily by individuals and businesses is rapidly increasing due to new technological advancements. The use of the Internet of Things and especially Machine-to-Machine communication channels have helped create a large interconnected network of computing devices. Specifically, the increasing use of mobile devices equipped with a large variety of sensors, various everyday embedded devices, and everyday tasks such as web browsing provide abundant information that is stored to the "cloud," i.e. to remote repositories. These data are then later accessed via Big Data infrastructures that propose methods to optimally extract, transform, and load this information to advanced systems capable of mining these data points. The outcome of these processes is to fuse the data streams in real-time and implement techniques harnessing the power of Artificial Intelligence (AI) to provide valuable data insights categorised into descriptive, diagnostic, predictive, and/or prescriptive results [1].

Additionally, AI is a field that shows great potential as, due to Moore's law that states the computing processing power yearly increases while the cost is decreased, we are now capable of handling rapidly and most importantly reliably data points generated on the spot. Analytically, AI is not a new term as one of the first mathematical models implementing a biological neural network (BNN) was first presented in 1943 [2]. This publication showcased how an Artificial Neural Network (ANN) can be emulated via the use of a BNN using advanced mathematic formulas consisting of parameters such as weights, bias, and activation functions [3], [4]. Furthermore, besides data classification tasks, this concept was later used as a stepping stone to enhance data insights in the field of Computer vision as it introduced detailed object classification and recognition in image/video datasets. Lastly, recent advances in the field of deep learning ANNS [5], [6], [7] focus on data classification and image recognition tasks via deep learning techniques [8], [9] in several fields. Most notably this includes e-commerce [10], finance [11], humanitarian aid [12], education [13], healthcare [14], and ecological informatics [15].

In this article we will focus on the last part of the above-mentioned process, presenting a detailed comparison of recent ML and deep learning techniques. Specifically, the outline of our paper is as follows: firstly, we will present the aims and objectives of our study. Secondly, we will briefly discuss some of the most widely used ML data and image classification techniques in the industry. Thirdly, we will present and focus on ML and Deep Learning solutions specialising in classification problems. Fourthly, we will present a detailed comparison analysis between these traditional ML and Deep Learning techniques. Fifthly, we will discuss our comparative results and suggest a novice ANN that achieves high accuracy of over 90% regarding a benchmark tested dataset. Finally, we will draw conclusions and discuss how this publication can be used in future works.

## 2. Aims and Objectives

This article aims to address the evaluation of some of the most widely used techniques regarding data and image classification. Specifically, our tests are two folded:

1. <u>ML techniques</u>: using the bag of visual words model (BOVW), K-nearest neighbours and support vector machine algorithms
2. <u>Deep Learning techniques</u>: using deep convolutional neural networks, i.e., a pre-trained model and the proposed ANN

Additionally, using the above-mentioned methods, we aim to address the following:

1. Provide information regarding some of the most widely used methods for data/image classification
2. Present a comparative study of ML and Deep Learning based solutions
3. Explain the architecture and the mathematical parameters of these methods
4. Suggest a novice CNN architecture for image classification

Lastly, the technical novelty of this article is not only presenting a comparative study between traditional ML and Deep Learning techniques, but suggesting a new CNN that achieves accuracy levels of slightly over 90% - and in some cases higher - similarly with the most recent scientific advances in the field.

## 3. Background and Methods

### 3.1 Defining the Problem

In machine vision, one of the most common problems mainly occur due to the following reasons: difficulties in recognising objects from different angles, differences in lighting, volatile rotation speeds of objects, rapid changes in scaling, and generic intraclass variations.

In the last decade, due to increasing computational power, the rise of cloud computing infrastructures, and advances in hardware acceleration techniques (either using GPUs or remote data centres), 2D object recognition research has rapidly increased. Specifically, recent research suggests that these challenges do not pose a computational problem [16], [17],

[18], [19], [20]. As a result, recent research has shifted its efforts to provide innovative solutions that take under consideration a trade-off between optimal accuracy for low-power and low-cost methods. This means that emphasis must be given to study, understand, and ameliorate existing computational mathematics approaches and methods.

In the following sections, we will present the 2 algorithms that we have studied using both ML and Deep Learning solutions. Analytically, the first focuses on ML extracting key features using SIFT algorithm and creating a global representation as described in the bag of visual words (BOVW) model [21], [22], [23]. Furthermore, we will present classification techniques focusing on the K-nearest neighbours' algorithm (KNN) and support vector machine (SVM) classifier. Moreover, regarding Deep Learning techniques we will showcase the implementation of two custom convolutional neural networks (CNN), i.e., one with and one without the use of a pretrained model. Lastly, for each of the method used we highlight the most important mathematic components and address common neural network issues such as overfitting.

### 3.2 Dataset

In order to conduct this comparative analysis, we have extended the BelgiumTS - Belgium Traffic Sign Dataset [24] by creating our custom version [25]. During our dataset design, we have noticed the existence of class imbalance containing more images than others, but this does not affect the quality of the data. Specifically, our version consists of several images from various traffic signs split into 34 classes with each class containing photos of different signs. More specifically, the training dataset contains 3056 images which are split into an 80/20 ratio regarding training and validation (i.e. 2457/599 images). Lastly, the testing dataset used contains 2149 images.

### 3.3 Traditional ML methods

The traditional ML methods that are most commonly used in academia and industry alike consist of the following:

1. Points of interest detection of points and features extraction (descriptions for each image of a trained data set).
2. Production of visual vocabulary based on the BOVW model and implementation of K-means algorithm.
3. Encoding training images using the generated dictionary and histogram extraction.
4. Classification using KNN and/or SVM

*1) Traditional ML methods*

In these methods, a system identifies the points of interests for each of the given images. Analytically, this is achieved via the use of detectors that enable features extraction (i.e., descriptions for a vector of features) for each examined point of interest. Then, these vectors are examined to determine the attribute identity thus enabling us to decide if two characteristics are similar. In this article, we use the SIFT descriptor [26].

*2) Production of visual vocabulary - BOVW model*

The BOVW consists of a dictionary, constructed by a clustering algorithm which aims to locate differences between an image and a general representation of a dataset. Specifically, the operating principle behind the BOVW model supports that to encode all the local features of an image, a universal representation of an image must be created. This model compares the examined image with the generated representation of each class and generates an output based on the differences of their content.

Similarly, in our article, our objective is to use an unsupervised learning technique that groups the output of all descriptors generated from an examined dataset of images to distinct groups of unique characteristics.

Furthermore, to implement this model, several algorithms approach, the most common one being the K-means, a clustering algorithm which organised the data points provided to the nearest centroid, for a fixed number K of clusters (i.e. words), until the system converged, for a given number of iterations [27]. The steps of this algorithms are the following [28]:

1. Initialise cluster centroids $\mu_1, \mu_2, \ldots, \mu_k \in R^n$ randomly
2. Repeat until convergence
   For every i, regarding the index of each point, set
   $$c^{(i)} := arg\ min_j \left\| x^{(i)} - \mu_j \right\|^2$$
   For every j, regarding the index of each cluster, set
   $$\mu_j := \frac{\sum_{i=1}^{m} \{c^{(i)} = j\} x^{(i)}}{\sum_{i=1}^{m} \{c^{(i)} = j\}}$$

Where, $x_i$ is the unique feature vector (descriptor) i and $c^{(i)}$ is the assigned cluster of xi.

### 3) Encoding

In addition, another step of great importance is to determine the properties of the classifier. Specifically, this is achieved via encoding the content of the images based on a dictionary of universal characteristics. In order to perform this, a histogram is produced that provides information regarding the frequency of the visual words of the dictionary in an image.

Moreover, upon producing a histogram for each word - using a vector of features - images are compared with a dictionary, and words correspond to the shortest distance. This results in finding the greatest similarity between the dataset.

Finally, we notice that normalisation is applied to the calculation of the occurrence frequency as we wished to ensure that the generated histograms did not depend on the number of visual words.

### 4) KNN classifier

The KNN algorithm is a non-parametric classifier [29], [30], [31] which accepts the histograms of the previous stage and compares them with the image dataset focusing on calculating and monitoring differences in the measured distances. Then, each image is classified to a unique cluster which shows the greatest degree of similarity with its k nearest neighbours.

As evident from the above, the classifier depends greatly on the distance metric used to predict and categorise each set of results into k-groups. Moreover, we should consider that a "one size fits all" solution does not exist and special attention should be given to each problem [32]. The distance measure selected highly depends on the dataset examined and should be chosen after a trial-and-error approach [33]. Specifically, many

distance measures exist, where the most commonly used are the following:

- **Manhattan Distance (L1)**, (also known as Taxicab), defines the distance ($d_1$), between 2 vectors (p, q) of an n-dimensional vector as follows:
  $$d_1(p,q) = \sum_{i=1}^{n} |p_i - q_i|$$

- **Euclidean Distance (L2)**, defines the distance ($d_2$) between 2 vectors (p, q) of an n-dimensional vector as follows:
  $$d_2(p,q) = \sqrt{\sum_{i=1}^{n} |p_i - q_i|^2}$$

Moreover, in machine learning, many distance metrics exist for multiple n-dimensional vector spaces to calculate the notion of distance or similarity [34]. In our study, we have selected to use a generalisation of L1 and L2 distances. Specifically, in our model we have defined the Minkowski distance, that based on the provided values of the input (p, q) it shapes its equation to the necessary distance. The Minkowski distance is defined as follows:

$$d_p(p,q) = \left( \sum_{i=1}^{n} |p_i - q_i|^p \right)^{\frac{1}{p}}$$

### 5) SVM classifier

The SVM classifier uses an algorithm to select the surface that lies equidistant from the 2 nearest vector spaces [35]. This is achieved via classifying dataset into different classes and calculating the margin between each class, thus creating vectors that support this margin region (support vectors). Additionally, in cases where several classes occur such as in our image recognition dataset, research also uses the "one versus all" approach where a classifier is trained for each class [36]. Analytically, the support vectors of the data points that are closer to the hyperplane influence the position and orientation of the hyperplane.

Furthermore, the mathematical equation for an SVM linear definition of a space (x) of examined data points is the following:

$$f(x) = w \cdot x + b$$

Moreover, the mathematical equation for a generic SVM definition is the following:

$$\{f : \|f\|_K^2 < \infty\}$$

Where, instead of a space (x), a space of a kernel K is used.

Lastly, based on the kernel space, the equations derived are the following:

$$K(x_1, x_2) = x_1 \cdot x_2, \quad f(x) = w \cdot x + b$$

Where the norm (w) of the equation is:

$$\|f\|_K^2 = \|w\|^2.$$

### 3.4 Deep Learning Algorithms

Apart from the ML methods discussed, Neural Networks (NN) are also used extensively combined with supervised learning techniques to identify and classify objects between classes [37]. Specifically, the most common use of these NN in the computer vision domain are CNNs implementing hierarchical feature learning [38]. These networks take advantage of the spatial information available on the ever-increasing areas of images during the system's processing. More specifically, due to their structure and their properties (e.g., Local Receptive Fields, Shared weights, Pooling) they significantly reduce the parameters and therefore the computational power over traditional feed-forward fully connected networks.

In our article, we have studied 2 models:
1. The VGG16 architecture [39] followed by a custom fully connected network as a classifier. We load the pretrained on Imagenet weights [40].
2. A custom CNN without transfer learning

Lastly, we notice that during our tests we have experimented with different activation functions of various hyperparameters, data augmentation and normalisation methods to avoid overfitting. In the following sections we will explain in detail each NN architecture as well as the activation functions and design principles.

### 6) VGG16 architecture

Firstly, we use the VGG16 architecture followed by our custom classifier. Specifically, we use fully connected layers with batch normalisation and dropout for regularisation, and mish activation function [41]. Lastly, the final layer of the classifier consists of 34 neurons using SoftMax activation functions, one for each class of the dataset.

Moreover, during each training phase, we maintain all the weights of each layer of VGG16 architecture frozen except for the last 4 layers. Specifically, we made this choice because the first convolutional layers "learn" low level features (such as straight lines, angles, edges, circles etc.) which are similar in most images. Although, the deeper convolutional layers "learn" high-level and more abstract features which are a combination of low-level features and are problem-specific to each dataset.

### 7) Proposed CNN architecture

The CNN architecture of the proposed Deep Learning architecture is presented in Figure 1 where the input dataset dimensions are: (128,128,3).

As evident from this image, the first stage of our CNN consists of a convolutional layer, where we perform BN and a max pooling. The second stage uses the same pattern twice, i.e., a convolutional layer followed by batch norm, another convolutional layer and BN and finally max pooling. Lastly, the classifier consists of 4 fully connected layers. The aim of the second stage is to achieve optimal performance as it is proposed for larger and deeper networks- as due to the multiple stacked convolutional layers, more complex features of the input volume can be extracted. This stage thus reduces cases of destructive pooling operation.

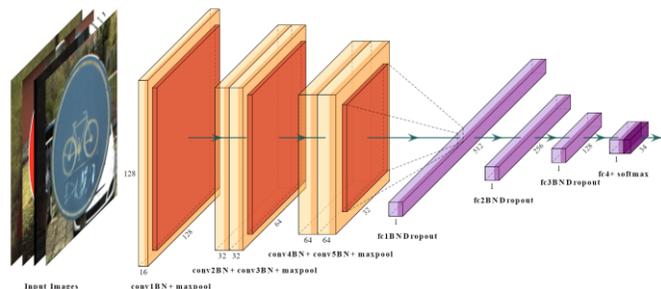

Fig.1. Architecture of the proposed NN where convolution, max pooling, and fully connected layers are highlighted in yellow, orange, and purple respectively

Furthermore, as the network progresses deeper, the number of applied filters is augmented. Specifically, initially we start with 16 filters and gradually increase them to 32 and 64. This increase assists in producing high quality features as it combines low-level features that occur while training the network. It is noted that in deep learning it is common practice to double the number of channels (or in our case filters) after each pooling layer as the knowledge of each layer becomes more spatial. As we get deeper into the network, each pooling layer divides the spatial dimensions by 2 thus doubling the number of filters and available kernels without the risk of rapid increases in parameters, memory usage, and computing load [42], [43], [44].

In addition, similarly to the above-mentioned architecture, this network also consists of 4 stages of fully connected layers where 3 of them use BN and dropout activation. Analytically, for this NN we have used several activation functions such as ReLU [45], [46], [47], and Swish [48] and after our tests, we conclude that Mish [49] is the optimal one. This function performed better than its rivals in terms of accuracy as it generated the smaller loss rates and a smoother loss landscape representation as its behaviour suggests that it avoids saturation due to capping. The Mish function is defined as follows:

$$f(x) = x \cdot tanh(softplus(x)) = x \cdot tanh(ln(1 + e^x))$$

Moreover, a fact of great importance is that smooth activation functions allow optimal information propagation deeper into the neural network, thus achieving higher accuracy and generalisation results. We present our findings regarding the activation functions studied during our experiments in Figure 2:

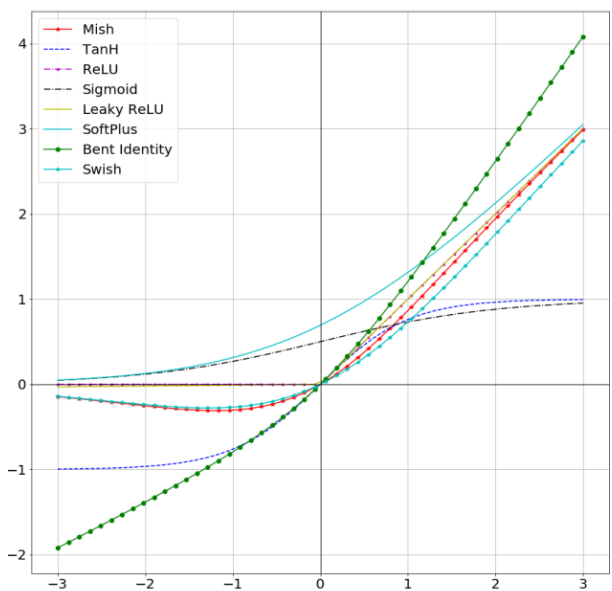

Fig.2. Comparison of the activation functions, some of them used in our experiments

Additionally, we have used adam optimiser [50] with a steady learning rate of 0.0002. Analytically, 80% of our examined dataset was used for training and 20% for validation. Moreover, to eliminate overfitting, we have used a BN technique similar to [51] and dropout technique.

The first was implemented by having continuously periodic measurements of the distributions for all the examined activations of each batch of our training examples at all levels. During the feed propagation phase, a normalisation of the output is applied for each batch, thus each batch average approximately reaches zero values and standard deviation reaches 1. Specifically, the mathematical equations used to compute these parameters are the following:

- Batch average:

$$\mu_B \leftarrow \frac{1}{m} \sum_{i=1}^{m} x_i$$

- Batch variation:

$$\sigma_B^2 \leftarrow \frac{1}{m} \sum_{i=1}^{m} (x_i - \mu_B)^2$$

- Normalised activation:

$$\hat{x}_i \leftarrow \frac{x_i - \mu_B}{\sqrt{\sigma_B^2 + \epsilon}}$$

- Changes in shift and scaling:

$$y_i \leftarrow \gamma \hat{x}_i + \beta \equiv \mathrm{BN}_{\gamma,\beta}(x_i)$$

Where $\gamma, \beta$ are the parameters to be learned by the trained NN and $\epsilon$ is constant for numerical stability that acts as a security check to eliminate the possibility denominator value is zero.

The second technique (i.e. dropout) used was similar to [52] where during the training of the NN a random value of neurons is selected to be disactivated (i.e. output 0). This acts as a means to force the NN to span out its usage and not deeply rely on specific neurons, thus generalising our solution and reducing the possibility of developing "single point of error" systems. This random value is selected based on a Bernoulli distribution, where, if p is the possibility of retaining activation and k of disactivating connections, NN behaviour is determined as:

$$\left(\frac{1}{p}\right)^k$$

Lastly, we used data augmentation methods, where training data are provided based on a probabilistic approach, i.e., calculating a possibility variable in real time during the training phase and transforming images on the fly. This technique was used during the training and not the testing or validation phase. This is because we need the NN to optimise its behaviour during learning and not execution mode.

## 4. Results

In this section we will firstly present the results of our network architecture using ML methods. Later, we will provide information regarding our experiments using Deep Learning methods and illustrate the advantages of our NN approach. Lastly, we will conclude this study with a comparative analysis consisting of confusion matrices and comparison diagrams of each architecture.

### 4.1 Traditional ML methods

The results of our executions are presented in Table 1 and 2 where Manhattan outperforms Euclidean on average by 10-15%. Specifically, we notice that while increasing the vocabulary size and thus the amount of vector space and their dimensions, this rate is rapidly increasing, similarly to recent studies in the bibliography [53]. Moreover, KNN classifier's optimal behaviour occurs for dictionaries of 75 words and the 50 nearest neighbours as well as of 100 words and 20 for Euclidian and Manhattan distance alike.

Lastly, as presented in Table 1 and 2 we notice that even though initially an increase in neighbours ameliorated the overall accuracy results, after a certain point, a threshold was reached as accuracy values stagnated or decreased. Moreover, as evident from Figures 3 and 4, results suggest that for a constant value of neighbours, there is a significant increase in performance if we increase the quantity of (visual) words (i.e., input). Unfortunately, this observation, although interesting, must be treated carefully as the initial rapid increase in accuracy applies only to small to medium-scale dictionaries. Specifically, for large dataset inputs, the trade-off between accuracy, computation cost, and execution time is not advisable similarly to the study of [54].

**Table 1.** Comparison of KNN accuracy for different values of K (Number of neighbours) and different small vocabulary sizes. Euclidian (L2) distance is used as distance metric

| K | Vocabulary Size | | |
|---|---|---|---|
| | **50** | **75** | **100** |
| 1 | 34.29502% | 37.22661% | 37.92462% |
| 3 | 33.73662% | 37.97115% | 38.20382% |
| 5 | 37.27315% | 38.34342% | 38.90181% |
| 10 | 38.34342% | 40.01861% | 40.29781% |

| K | Vocabulary Size | | |
|---|---|---|---|
| | *50* | *75* | *100* |
| 15 | 40.01861% | 41.13541% | 41.18195% |
| 20 | 39.13448% | 40.67008% | 41.55421% |
| 50 | 39.69288% | 42.57794% | 41.27501% |
| 75 | 38.94835% | 42.34528% | 41.55421% |
| 100 | 39.46021% | 41.74034% | 41.78688% |

**Table 2.** Comparison of KNN accuracy for different values of K (Number of neighbours) and different small vocabulary sizes. Manhattan (L1) distance is used as distance metric.

| K | Vocabulary Size | | |
|---|---|---|---|
| | *50* | *75* | *100* |
| 1 | 41.32154% | 49.55793% | 52.44300% |
| 3 | 42.67101% | 50.58167% | 53.93206% |
| 5 | 46.25407% | 52.62913% | 54.11819% |
| 10 | 48.39460% | 55.00233% | 58.53886% |
| 15 | 47.41740% | 54.76966% | 58.49232% |
| 20 | 48.44114% | 55.56073% | 58.49232% |
| 50 | 48.58074% | 53.88553% | 57.09632% |
| 75 | 47.27780% | 52.81526% | 55.93299% |
| 100 | 46.39367% | 51.60540% | 54.49046% |

**Table 3.** Comparison of SVM accuracy for different types of Kernels and different small vocabulary sizes.

| Kernel | Vocabulary Size | | |
|---|---|---|---|
| | *50* | *75* | *100* |
| RBF | 30.71196% | 36.57515% | 43.78781% |
| LINEAR | 28.05956% | 42.29874% | 44.95114% |
| SIGMOID | 4.18799% | 4.18799% | 4.18799% |
| CHI2 | 48.90647% | 57.18939% | 54.25779% |
| INTER | 47.83620% | 52.81526% | 56.11913% |

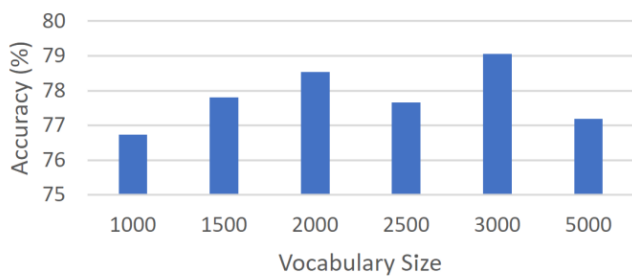

Fig.3. Accuracy comparison of kernel SVM kernel for increasing number of vocabulary sizes.

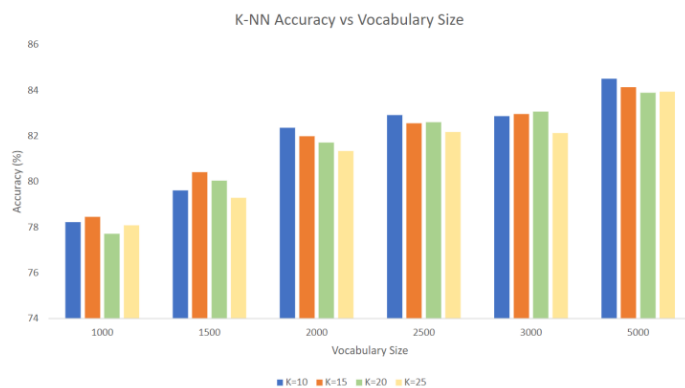

Fig.4. Accuracy comparison of KNN for various k clusters and an increasing number of vocabulary sizes.

## 4.2 AI methods

As for the examined AI methods, we have used call backs and early stopping aiming to halt the training phase if a metric defined by the validation loss parameter reduces its optimisation (i.e., value decrease) for a preset number of epochs. Analytically, the training phase of our proposed NN model peaked its accuracy at epoch 35 (optimal validation loss value) and after 12 epochs its training stopped and retrieved the previous weights of epoch 35. We will present our findings regarding the accuracy and the loss curves for our proposed solution - in comparison to the pretrained model used as a means of reference - in Figure 5 and 6.

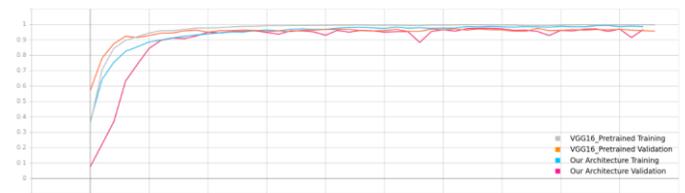

Fig.5. Accuracy curves regarding the proposed model and the pretrained model during training and validation

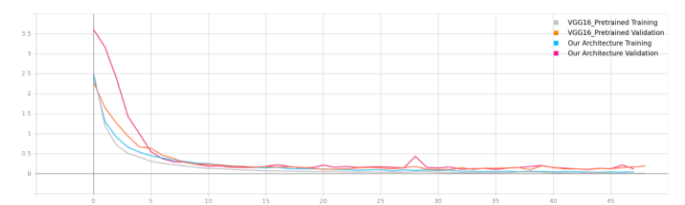

Fig.6. Loss curves regarding the proposed model and the pretrained model during training and validation.

## 4.3 Comparative Analysis

Except for comparing the dictionary size and focusing on either the BOVW or the KNN classifier, we have also studied the behaviour of our suggested system with traditional metrics used to access NN architectures. Firstly, we will present in Table 4 the accuracy of our tests. Secondly, we will introduce the confusion matrices regarding the optimal results for each case. Specifically, we will provide a detailed visualisation of our comparative study in Figures 7, 8, and 9 for all of the above-mentioned methods.

**Table 4.** Training Accuracy and Loss between Proposed and pretrained model (VGG16)

| Benchmark Metrics | | Proposed | Pretrained (VGG16) |
|---|---|---|---|
| Training | Accuracy | 0.9833 | 0.9897 |
| | Loss | 0.0616 | 0.0597 |
| Validation | Accuracy | 0.9716 | 0.9750 |
| | Loss | 0.1075 | 0.1011 |
| Testing | Accuracy | 0.9744 | 0.9716 |
| | Loss | 0.1321 | 0.1191 |

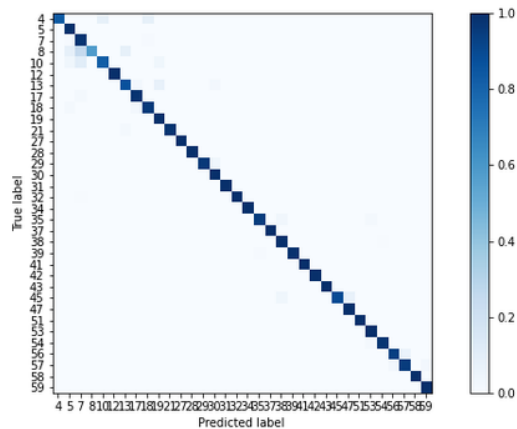

Fig.9. Confusion matrix of the proposed DCNN.

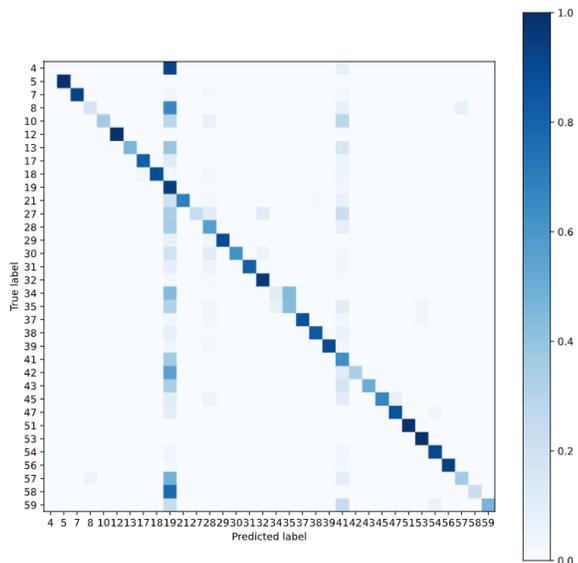

Fig.7. Confusion matrix of the optimal SVM (Kernel= inter, Vocab_size = 3000).

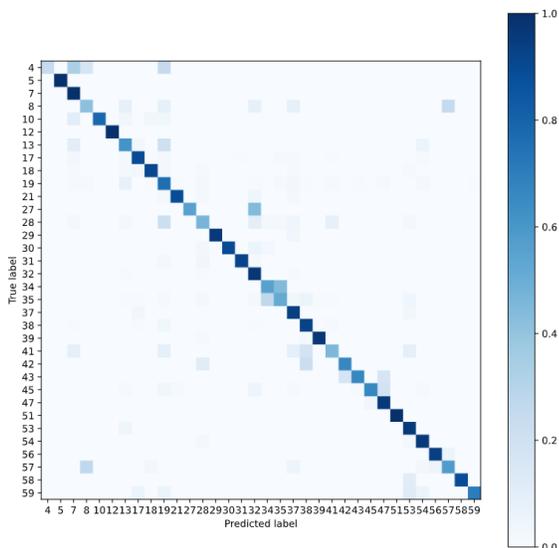

Fig.8. Confusion matrix of the optimal KNN ( K= 10, Vocab_size = 5000).

# 5. Conclusion and Future Works

In this article, we presented in detail a novice architecture of a CNN model with validation and testing loss of 0.1075 and 0.1321 alike. Moreover, the system presented had a high validation and testing accuracy of 97.16% and 97.44% respectively, achieving optimal results in various test cases. In addition, the pretrained model used as a point of reference and comparison with our suggested CNN, initially scored lower loss and higher validation accuracy (57% after the $1^{st}$ epoch) and finally reached validation and testing loss of 0.1011 and 0.1191 and validation and testing accuracy of 97.5% and 97.16% alike.

The comparison between the proposed and pretrained model shows that our system is capable of achieving similar results with existing CNNs but authors suggest that for medium to small size datasets, it slightly outperforms existing CNN architectures. Analytically, authors suggest that this study can be used in self-driving vehicle navigation or routing systems [55], [56], [57] for object detection and avoidance. Also, since our system was designed to use low-power and low-computational cost hardware properties, we suggest this CNN to be used either as a standalone solution or as part of existing vehicles' automatic control systems used to validate the prediction (i.e. ground truth) of other sensory data points (e.g. from camera or lidar) predictions.

Finally, future system expansions should shift their attention into developing a dynamic model that automatically searches and combines existing activation functions to achieve better results. Moreover, we need to expand our architecture similarly to the studies of vision transformers designs, studying related datasets such as [58], [59], [60] as well as self-supervised pretraining techniques [61].